\pdfoutput=1

\documentclass[11pt]{article}

\usepackage{acl}

\usepackage{times}
\usepackage{latexsym}
\usepackage{graphicx}
\usepackage{pgfplots}
\usepackage{filecontents}
\usepackage{float}
\usepackage{hyperref}

\usepackage[T1]{fontenc}

\usepackage[utf8]{inputenc}

\usepackage{microtype}

\title{AIFB-WebScience at SemEval-2022 Task 12: Relation Extraction First - Using Relation Extraction to Identify Entities}

\author{
  \begin{tabular}{c}
  Nicholas Popovic \\
  \textnormal{Karlsruhe Institute of Technology (KIT), Germany} \\
  {\tt popovic@kit.edu} 
   \end{tabular}%
   \begin{tabular}{c}
  Walter Laurito\\
  \textnormal{FZI Research Center for Information Technology, Germany} \\
  {\tt laurito@fzi.de} 
   \end{tabular} \\[1.5em]
  \begin{tabular}{c}
  \textbf{Michael Färber} \\
  \textnormal{Karlsruhe Institute of Technology (KIT), Germany} \\
  {\tt michael.faerber@kit.edu} 
   \end{tabular}%
   }

\begin{document}
\maketitle
\begin{abstract}
In this paper, we present an end-to-end joint entity and relation extraction approach based on transformer-based language models.
We apply the model to the task of linking mathematical symbols to their descriptions in LaTeX documents.
In contrast to existing approaches, which perform entity and relation extraction in sequence, our system incorporates information from relation extraction into entity extraction.
This means that the system can be trained even on data sets where only a subset of all valid entity spans is annotated.
We provide an extensive evaluation of the proposed system and its strengths and weaknesses.
Our approach, which can be scaled dynamically in computational complexity at inference time, produces predictions with high precision and reaches 3rd place in the leaderboard of SemEval-2022 Task 12.
For inputs in the domain of physics and math, it achieves high relation extraction macro $F_1$ scores of 95.43\% and 79.17\%, respectively.
The code used for training and evaluating our models is available on GitHub\footnote{\url{https://github.com/nicpopovic/RE1st}}. 
\end{abstract}

\section{Introduction}

Information extraction systems are a key component in making scientific literature more consumable.
With the large amount of scientific works which are constantly being published (e.g., more than 60,000 machine learning papers per year \cite{DBLP:conf/semweb/Farber19}), indexing techniques that go beyond keyword searches are becoming more important. 
While many efforts have focused on the processing of abstracts as a way of building representations of publications \cite{gabor_semeval-2018_2018, luan_multi-task_2018}, methods processing full text documents will be needed to accurately capture their contents for use cases such as academic search and recommender systems and scientific impact quantification.

The task tackled in this paper \cite{lai-etal-2022-semeval}, consisting of linking mathematical symbols to their descriptions in LaTeX documents, is a \textit{joint entity and relation extraction} task.
While earlier work tackled both subtasks sequentially via separate models, more recent approaches tend to use a single joint model \cite{luan_multi-task_2018, bekoulis_joint_2018, nguyen_end--end_2019, eberts_end--end_2021}.
In contrast to early approaches, which are based on Bi-LSTMs \cite{luan_multi-task_2018, bekoulis_joint_2018, nguyen_end--end_2019}, more recent approaches \cite{wadden_entity_2019, eberts_end--end_2021} make use of transformer-based language models, such as BERT \cite{devlin_bert_2019}.
A key challenge in joint models is the computational complexity stemming from pairwise comparisons between entity spans required for relation extraction.
Previous works tackle this using a span scoring mechanism based on a feed forward neural network, which produces a score indicating the likelihood that a span is in a relation \cite{luan_multi-task_2018, wadden_entity_2019}.
Relation extraction is then performed on only those spans with the highest scores.
For data sets which include span annotations even for entities which are not in any relation, such as DocRED \cite{yao_docred_2019}, as examined by  \citet{eberts_end--end_2021}, such a scoring mechanism is not necessary, because the entity extraction component of the model can be trained on these annotations.
For the task tackled in this paper, complete annotations for entity spans are not provided, making the use of a span scoring mechanism necessary.

\begin{figure*}
    \centering
    \includegraphics[width=1\textwidth, angle=0]{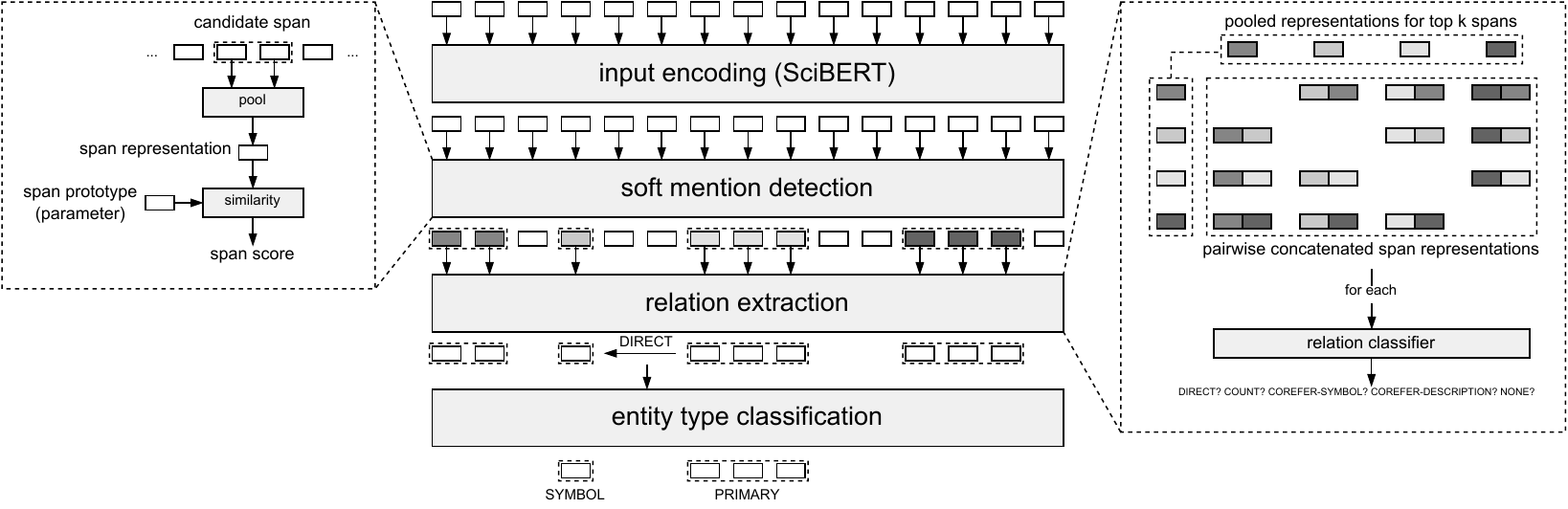}
    \caption{Architecture overview with detail illustrations for the soft mention detection (left) and relation extraction (right) modules. The layout of this figure was inspired by a similar figure found in \cite{eberts_end--end_2021}.}
    \label{fig:overview}
\end{figure*}

In this paper, we propose an end-to-end approach for joint entity and relation extraction. The approach is based on a transformer-based language model, following previous work \cite{DBLP:conf/ecai/EbertsU20, eberts_end--end_2021}, but is peculiar in the sense that it incorporates a span scoring mechanism based on dot product similarity which is learned via triplet loss rather than cross entropy loss, making it applicable to datasets which contain annotations only for a subset of all valid entity mention spans.

\section{Task Description}

The task tackled in this paper is one of joint entity and relation extraction. This means, given an unannotated text as input, a system needs to (1)~return annotations of relevant entity mention spans, (2)~perform coreference resolution, (3)~entity type classification, and finally (4)~relation extraction on the identified spans.
The specific task at hand has a number of key features that separate it from similar 
settings. 

First, regarding entity extraction, the annotations and, thus, the final scoring are restricted to those entities which participate in relations. 
This means that a system which correctly identifies all symbols and descriptions in the input will score poorly even on the entity extraction portion of the final benchmark if the relation extraction is incorrect.
More importantly from an engineering perspective, the resulting span annotations are incomplete in that they only include a partial set of valid spans for each document. 
In the entity extraction step we can, therefore, only reliably identify true positives and false negatives, not, however, false positives and true negatives.

Second, while coreference resolution (i.e., the linking of multiple mentions to a single entity) is part of the task, relation extraction is to be performed on a mention-level rather than the entity-level.
This means that although a system may correctly identify a text span as being the description of a certain symbol, this classification will only be deemed correct in the evaluation if linked to the correct mention of said symbol. 
As a result, coreference links are interpreted as relations between mentions and thereby as part of the relation extraction subtask, rather than as part of the entity extraction subtask.

Third, entity types can be reliably inferred from the relations between them, meaning that instances of relations are only found between certain entity types. 
This feature can be used to inform the design of a system in two ways: Either, the task of relation extraction can be simplified by reducing the choices given to a classifier based on the entity types of two spans (i.e., a symbol cannot be the description to another symbol, therefore any such prediction can be disregarded), or the entity type classification can be informed by the relation extraction (i.e., if we identify a span $A$ as the description of another span $B$, span $A$ must be a description, while span $B$ must be a symbol).

\section{Approach}

We propose an end-to-end entity and relation extraction system using a transformer-based language model, as illustrated in figure \ref{fig:overview}. 
The system consists of 4 modules: (1) The \textit{input encoding module} tokenizes the input text and produces contextualized embeddings for each token, (2) the \textit{soft mention detection module} ranks possible token spans by the likelihood with which they contain an entity mention, (3) the \textit{relation extraction module} extracts relations on a subset of the highest ranked spans from the previous step, and finally (4) the \textit{entity type classification module} assigns entity types to spans based on the relations detected between them.

\subsection{Input Encoding}
\label{sec:preprocessing}
We examine two separate options of encoding the input: 
For the first option, we pass the input text to the language
model without prior modification, whereas for the second option, we
perform preprocessing on the input to remove LaTeX code from the text portions of the input. Any
input in LaTeX math mode is passed to the model
unchanged.

Since our approach uses a transformer-based language model, the input needs to be tokenized. %
As a result of the tokenization, there are instances of relations which cannot be matched correctly by our model, due to the annotated span boundaries being contained within a token. 
For the training and development sets, this occurs in 1.99\% and 2.84\% of relation instances, respectively, and in these cases we adjust the labels accordingly.

\subsection{Soft Mention Detection}
\label{sec:mention_localization}
Given that we cannot reliably identify false positives and true negatives from our labeled data, a mention detection strategy based on cross-entropy loss cannot be used for this task.
Instead of following previous approaches in using feed-forward neural networks \cite{luan_multi-task_2018, wadden_entity_2019}, we propose a linear similarity based approach which ranks possible spans based on their similarity to multiple \textit{prototype} embeddings (one prototype per entity type).

We begin by computing the set of all possible continuous spans up to a maximum length $n$ and produce a fixed-size embedding $e_{s}$ for each span by pooling the contextualized embeddings of all tokens within it.
As pooling strategies we use either mean or max pooling.
For each span embedding $e_{s}$ we compute a span score $X_{s}$:
\begin{equation}
    X_{s} = \max_{a_{i} \in A} (sim(e_{s}, a_{i}))
\end{equation}
where $A$ is the set of prototype embeddings which contains an embedding for each entity type and $sim(a,b)$ is the dot product similarity of two vectors.
We select the $k$ spans with the highest values for $X_{s}$ as our \textit{candidate mentions} $M$ for relation extraction. 
We compute the \textit{mention loss} as the mean triplet loss \cite{DBLP:conf/cvpr/SchroffKP15} across all prototype embeddings in $A$ and all mentions in $M$.

\subsection{Relation Extraction}

For relation extraction, we use the document-level relation extraction model DL-MNAV \cite{popovic_fsdlre_2022}. 
We use the concatenation of two span representations as a representation for the relation between them \cite{wang_extracting_2019}.
The resulting relation representations are compared to a single relation prototype embedding per relation type, as well as $m$ additional prototypes representing the none-of-the-above class (this follows the MNAV model \cite{sabo_revisiting_2021}).
The relation type corresponding to the prototype resulting in the highest dot product similarity for a relation representation is used as the predicted type.
As loss function for the relation classification we use \textit{adaptive thresholding loss} \cite{zhou_document-level_2020} as it is capable of handling the large imbalance between positive and negative training examples present in document-level relation extraction tasks.\\

Due to quadratic scaling of the pairwise comparisons it is not feasible to perform relation extraction on all possible continuous spans.
We, therefore, perform relation classification on the top $k$ spans\footnote{During training we add annotated spans which are not among the top $k$ spans.} with the highest span scores, meaning that we have to classify a maximum of $k(k-1)$ relation representations for a given input text.
The computational complexity of the system can, therefore, be adjusted dynamically at inference time by changing $k$, for example to be run on GPUs with smaller memory capacity or on GPUs with higher memory capacity to improve the quality of predictions. \\

As a result of the soft mention detection, it is possible that some of the $k$ spans are overlapping and correspond to the same target (see appendix \ref{sec:appendix_b} for examples).
This means that the relation classifier may output multiple predictions for the same relation instance with slightly different mention spans. 
For predictions in which both the head and tail entity overlap, we therefore output only the prediction with the highest classification score.

\subsection{Entity Type Classification}

Finally, we use a simple mapping to determine the entity type of the spans which participate in the relations predicted by the relation classifier. 
The mapping used can be found in appendix \ref{sec:appendix_a}. For spans classified as "PRIMARY" we additionally change the predicted type to "ORDERED", if they are the head entity of more than one "Direct" relation.

\begin{table*}
\centering
\scalebox{0.7}{
\begin{tabular}{ll|cccc|ccc}
\hline
&&\multicolumn{4}{c|}{Entity Extraction}&\multicolumn{3}{c}{Relation Extraction}\\
\hline
pooling & preprocessing & $F_1$ strict [\%] & $F_1$ exact [\%] & $F_1$ partial [\%] & $F_1$ type [\%] & precision [\%] & recall [\%] & $F_1$ [\%] \\
\hline
\multicolumn{9}{c}{Development set}\\
\hline
max & None & \textbf{59.79 $\pm$ 0.99} & \textbf{60.19 $\pm$ 0.99} & \textbf{69.20 $\pm$ 0.68} & \textbf{67.45 $\pm$ 1.05} & \textbf{65.86 $\pm$ 1.34} & 44.14 $\pm$ 0.97 & \textbf{52.86 $\pm$ 1.10} \\
mean & None & 58.02 $\pm$ 3.67 & 58.49 $\pm$ 3.70 & 69.14 $\pm$ 2.02 & 66.98 $\pm$ 2.31 & 61.27 $\pm$ 5.09 & \textbf{44.58 $\pm$ 1.56} & 51.61 $\pm$ 2.87 \\
max & LaTeX2Text & 58.90 $\pm$ 0.79 & 59.30 $\pm$ 0.87 & 68.69 $\pm$ 1.00 & 66.88 $\pm$ 1.09 & 64.54 $\pm$ 2.61 & 43.77 $\pm$ 1.76 & 51.66 $\pm$ 0.77 \\
mean & LaTeX2Text & 54.59 $\pm$ 15.07 & 54.99 $\pm$ 14.44 & 65.62 $\pm$ 11.28 & 64.22 $\pm$ 14.22 & 63.89 $\pm$ 33.24 & 40.59 $\pm$ 15.90 & 49.64 $\pm$ 23.63 \\
\hline
\multicolumn{9}{c}{Test set}\\
\hline
max & None & - & - & 37.83 $\pm$ 0.85 & 37.88 $\pm$ 0.85 & 45.80 $\pm$ 5.80 & 20.96 $\pm$ 0.08 & 28.66 $\pm$ 1.19\\
mean & None & - & - & \textbf{41.21 $\pm$ 1.18} & \textbf{41.23 $\pm$ 1.19} & 42.25 $\pm$ 3.19& \textbf{26.55 $\pm$ 1.19} & \textbf{32.28 $\pm$ 0.20} \\
max & LaTeX2Text & - & - & 38.33 $\pm$ 1.57 & 38.38 $\pm$ 1.57 & 46.09 $\pm$ 0.77 & 21.64 $\pm$ 1.60 & 29.45 $\pm$ 1.41 \\
mean & LaTeX2Text & - & - & 34.53 $\pm$ 11.02 & 34.64 $\pm$ 11.13 & \textbf{47.02 $\pm$ 20.70} & 18.20 $\pm$ 8.10 & 26.24 $\pm$ 11.64 \\
\hline
\end{tabular}
}
\caption{\label{tab:results_test} Entity and relation extraction scores for 4 different models on both the development and the test set. NER metrics strict and exact were not produced by the test set evaluation script on the competition site and the test set is not publicly available at the time of writing.}
\end{table*}

\section{Experimental Setup}

For our language model we use SciBERT \cite{beltagy_scibert_2019}, which is trained on scientific text, via Huggingface's Transformers library \cite{wolf_transformers_2020}. 
For LaTeX preprocessing (see section \ref{sec:preprocessing}) we use Pylatexenc\footnote{\url{https://github.com/phfaist/pylatexenc}}. 
As our optimizer, we use AdamW \cite{loshchilov_decoupled_2019} with learning rates $\in [3\mathrm{e}{-5}, 5\mathrm{e}{-5}, 7\mathrm{e}{-5}]$, a linear warmup of 1 epoch followed by a linear decay to zero, for a total of 60 epochs\footnote{The length of one epoch is dictated by the number of training examples, which is 3119.}, a batch size of 4, and apply gradient clipping with a max norm of 1. 
During training, we randomly downsample the amount of candidate spans for soft mention detection to 1000, while ensuring that all labeled spans are included. During training and development set evaluation, we set $k$, the number of spans to perform relation classification on, to 50, as preliminary experiments showed this value to yield a good compromise between model performance and training time. For test set evaluation we increase $k$ to 400.
Training takes approximately 10 hours on a single NVIDIA V100 GPU using mixed precision. 
We perform early stopping based on the micro $F_1$ score for relation extraction on the development set. 
We train each hyperparameter configuration 3 times using different random seeds and report the median and standard deviation for each metric. 
As a result of the different combinations of preprocessing and mean-/max-pooling, we examine the performance of 4 configurations on the test set.
For our evaluation, we report the micro $F_1$ scores for NER metrics as used in SemEval-2013 Task 9.1 \cite{segura-bedmar_semeval-2013_2013}\footnote{We use the following implementation: \url{https://github.com/davidsbatista/NER-Evaluation}}.
For relation extraction we report \textit{micro} precision, recall and $F_1$ scores, unless otherwise indicated.
\section{Results}

\subsection{Overview}

The results of the 4 model configurations on the test set are reported in table \ref{tab:results_test}.
In comparison to the other approaches taking part in SemEval-2022 Task 12, our system ranks in place 3/9 in terms of relation extraction $F_1$ score.\footnote{Scores for other metrics are not publicly visible on the leaderboard at the time of writing.}

In general, we find that our model produces predictions with significantly higher precision than recall.

\subsection{Impact of Preprocessing}
With respect to the preprocessing procedure, we observe no clear performance impact.
We conclude that SciBERT appears to cope well with LaTeX code and preprocessing, as described in this paper, is not required.
\subsection{Impact of Pooling Procedure}
Regarding the pooling procedures we find that mean pooling tends to cause higher variability in the classification performance of the models.
For the models trained using mean pooling and preprocessing, 1 of 3 models performed significantly worse than the others, causing the large standard deviation in the results.

\subsection{Impact of Domain}

\begin{table}
\centering
\scalebox{0.8}{

\begin{tabular}{lrrrr}
\hline
 & \multicolumn{4}{c}{domain} \\
\hline
 & cs & econ & math & physics \\
\hline
\% of training corpus & 16.63 & 27.08 & 12.82 & 32.08 \\
\hline
relation type &  &  &  &  \\
\hline
Direct & 33.12 & 21.05 & 63.82 & 85.71 \\
Count & - & - & 84.62 & 100.00 \\
Corefer-Symbol & 21.05 & 20.47 & 91.30 & 100.00 \\
Corefer-Description & 3.51 & 0.00 & 76.92 & 96.00 \\
\hline
macro & 19.23 & 13.84 & 79.17 & 95.43 \\
micro & 25.93 & 19.49 & 78.77 & 88.77 \\

\hline
\end{tabular}
}
\caption{\label{tab:domain_re} F1 scores for relation extraction across different domains and relation types on the development set. cs and econ do not contain any instances of "Count".}
\end{table}

In table \ref{tab:domain_re}, we show the relation extraction $F_1$ scores for a model across the 4 different domains covered by the development set paired with the distribution of training data across domains.
We observe large performance differences depending on the domain with math and physics showing very high macro $F_1$ scores (79.17\% / 95.43\%) and computer science and economics performing poorly (19.23\% / 13.84\%).
While physics content does represent the majority of training examples, the distribution of domains across training examples does not fully explain the disparity.

\subsection{Impact of $k$}

\begin{figure}
    \centering
    \scalebox{0.83}{
    \begin{tikzpicture}
    \begin{axis}[legend pos=south east, ylabel=\%, xlabel=$k$]
    \addplot[color=black,smooth, thick, dotted] table [x=x, y=p, col sep=comma] {data.csv};\addlegendentry{RE precision}
    \addplot[color=black,smooth, thick, dashed] table [x=x, y=r, col sep=comma] {data.csv};\addlegendentry{RE recall}
    \addplot[color=black,smooth, thick] table [x=x, y=f, col sep=comma] {data.csv};\addlegendentry{RE $F_1$}
    \addplot[color=blue,smooth, thick, dashed] table [x=x, y=re, col sep=comma] {data.csv};\addlegendentry{entity recall}
    \end{axis}
    \end{tikzpicture}
    }
    \caption{Plot of the impact of increasing values of $k$ on precision, recall, and F1 scores on the development set.}
    \label{fig:k_rel}
\end{figure}
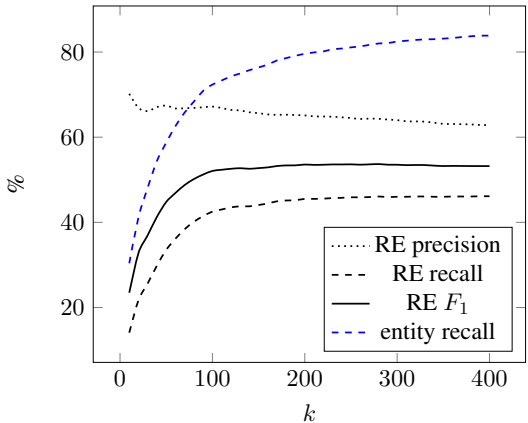

In figure \ref{fig:k_rel}, we show the change in relation extraction performance across different values for $k$.
We also include in the plot the percentage of entity spans in the top $k$ ranked spans (entity recall).
While the relation extraction performance improves proportional to the entity recall for $k \leq 100$ the improvement slows down for higher $k$.
We hypothesize that this is due to the limiting of $k=50$ and the candidate span downsampling during training, which prevents the model from seeing some of the more difficult cases.
In appendix \ref{sec:appendix_b}, we show examples of detected spans.

\subsection{Impact of Tokenization}

\begin{table}
\centering
\scalebox{1}{
\begin{tabular}{lrrr}
\hline
matching & precision & recall & $F_{1, micro}$ \\
\hline
strict & 55.85 & 44.01 & 49.23 \\
partial & 62.87 & 46.13 & 53.22 \\
\hline
\end{tabular}
}
\caption{\label{tab:partial_matching} Comparison of strict and partial matching requirements with respect to classification scores on the development set.}
\end{table}

In order to measure the impact of tokenization errors produced by adjusting labels during training, we perform a partial matching of relation labels as follows:
For predicted relation triples which are false positives, we accept them as true positives for an annotated instance if the \textit{intersection-over-union} (IOU) scores of both head and tail entities are greater than 67\% and the predicted relation type matches the label. %
In table \ref{tab:partial_matching} we show the results of both strict and partial matching for our best model on the development set.
We find that the relaxed requirements for span accuracy result in an increase in the $F_1$ score of 3.99\%. 
We conclude that tokenization errors, while measurable, do not account for the majority of errors of our model.

\section{Conclusion}

In this paper, we present an end-to-end joint entity and relation extraction approach for linking mathematical symbols to their descriptions in LaTeX documents.
Our model appears to be sensitive to the domain of the input documents, achieving high macro $F_1$ scores of 95.43\% and 79.17\% for physics and math content, respectively, while achieving macro $F_1$ scores of only 19.23\% and 13.84\% for computer science and economics related content.
We find that the model's predictions are higher in precision than in recall.
We perform a detailed error analysis and identify cross-domain generalization as the most critical problem to tackle in future work.

\section*{Acknowledgements}

The authors acknowledge support by the state of Baden-Württemberg through bwHPC.

\bibliography{references}
\bibliographystyle{acl_natbib}

\appendix

\section{Appendix}
\label{sec:appendix}

\subsection{Entity Type Classification Map}
\label{sec:appendix_a}

Table \ref{tab:rules} shows the classification map used for determining entity types based on relations for SemEval-2022 Task 12.

\begin{table}[h]
\centering
\scalebox{0.9}{
\begin{tabular}{lcc}
\hline
\textbf{Relation} & \textbf{head entity} & \textbf{tail entity}\\
\hline
Direct & PRIMARY* & SYMBOL\\
Count & PRIMARY & SYMBOL\\
Corefer-Symbol & SYMBOL & SYMBOL\\
Corefer-Description & PRIMARY & PRIMARY\\
\hline
\end{tabular}
}
\caption{\label{tab:rules}Classification map for entity types based on relations in which the spans participate. *In a postprocessing step, entity types of spans which are the head entity of multiple "Direct" relations are adjusted to "ORDERED".}
\end{table}

\subsection{Examples of Spans Detected for Different Values of $k$}
\label{sec:appendix_b}
\begin{figure*}
    \centering
    \includegraphics[width=1\textwidth, angle=0]{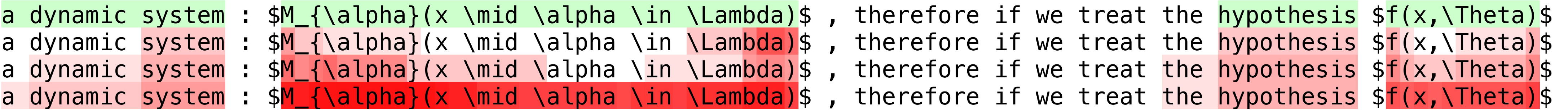}
    \caption{An example of spans detected in the domain of computer science. The top row shows ground truth labels in green, while the rows below are spans detected at $k=50,100,150$.}
    \label{fig:spans_cs}
\end{figure*}

\begin{figure*}
    \centering
    \includegraphics[width=1\textwidth, angle=0]{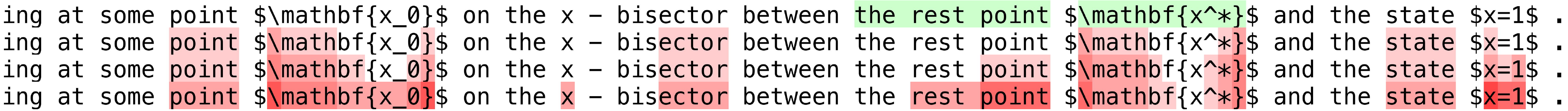}
    \caption{An example of spans detected in the domain of economics. The top row shows ground truth labels in green, while the rows below are spans detected at $k=50,100,150$.}
    \label{fig:spans_econ}
\end{figure*}

\begin{figure*}
    \centering
    \includegraphics[width=1\textwidth, angle=0]{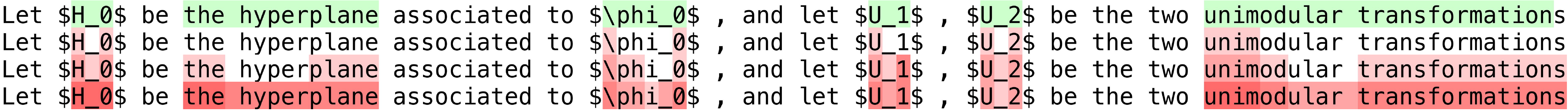}
    \caption{An example of spans detected in the domain of mathematics. The top row shows ground truth labels in green, while the rows below are spans detected at $k=50,100,150$.}
    \label{fig:spans_math}
\end{figure*}

\begin{figure*}
    \centering
    \includegraphics[width=1\textwidth, angle=0]{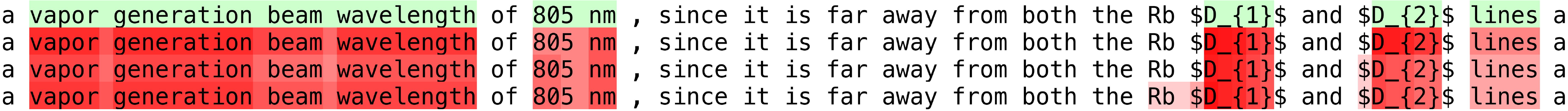}
    \caption{An example of spans detected in the domain of physics. The top row shows ground truth labels in green, while the rows below are spans detected at $k=50,100,150$.}
    \label{fig:spans_physics}
\end{figure*}

Examples of spans detected via soft mention detection are shown in figures \ref{fig:spans_cs}, \ref{fig:spans_econ}, \ref{fig:spans_math}, and \ref{fig:spans_physics}.


\end{document}